# MegaChat: A Synthetic Persian Q&A Dataset for High-Quality Sales Chatbot Evaluation

Mahdi Rahmani
*Agentic AI Research Department*
*Eastern Smart Innovators*
Tehran, Iran
Mahdi@MegaChat.ir

AmirHossein Saffari
*Agentic AI Research Department*
*Eastern Smart Innovators*
Tehran, Iran
AmirHossein@MegaChat.ir

Reyhane Rahmani
*Agentic AI Research Department*
*Eastern Smart Innovators*
Tehran, Iran
Reyhane@MegaChat.ir

*Abstract*— Small and medium-sized enterprises (SMEs) in Iran increasingly leverage Telegram for sales, where real-time engagement is essential for conversion. However, developing AI-driven chatbots for this purpose requires large, high-quality question-and-answer (Q&A) datasets, which are typically expensive and resource-intensive to produce, especially for low-resource languages like Persian. In this paper, we introduce MegaChat, the first fully synthetic Persian Q&A dataset designed to evaluate intelligent sales chatbots in Telegram-based e-commerce. We propose a novel, automated multi-agent architecture that generates persona-aware Q&A pairs by collecting data from active Telegram shopping channels. The system employs specialized agents for question generation, validation, and refinement, ensuring the production of realistic and diverse conversational data. To evaluate answer generation, we compare three classic retrieval-augmented generation (RAG) models with our advanced agentic system, which features multi-query retrieval, reranking, and persona-aligned response synthesis. Using GPT-5.1 for evaluation across six quality dimensions, our results show that the agentic architecture outperformed traditional RAG models in 4 out of 5 diverse channels, demonstrating its ability to generate scalable, high-quality datasets without relying on expensive human annotation or complex fine-tuning. MegaChat provides SMEs with an efficient, cost-effective solution for building intelligent customer engagement systems in specialized commercial domains, enabling advancements in multilingual conversational AI for low-resource languages.

Download: https://github.com/MegaChat-Tech/MegaChat-DataSet

## I. INTRODUCTION

### A. SMEs Driving Global Innovation and Economic Growth

Small and medium-sized enterprises (SMEs) are the backbone of global innovation and employment. They account for 70% of jobs worldwide and represent over 90% of businesses in the U.S., while in China they dominate the enterprise landscape, contributing significantly to GDP, taxation, and innovation [1], [2], [3]. Their scale and diversity make them critical drivers of economic resilience and technological progress.

### B. Digital Platforms Empowering SMEs

Digital platforms have transformed SME marketing and customer engagement. Social media channels such as Facebook, Instagram, and Twitter allow firms to reach wide audiences, build long-term relationships, and operate cost-effectively [4]. In Iran, Telegram has emerged as a uniquely dominant platform, with over 900 million monthly active users globally. Its low advertising costs, accessible design, and support for channels, groups, and bot-based services make it a central tool for SME sales and communication [5], [6], [7].

### C. Real-Time Engagement and AI Chatbots

Customer expectations increasingly demand immediacy: leads contacted within five minutes are 21 times more likely to convert, and timely messaging boosts engagement by up to 40% [8], [9]. AI chatbots address this challenge by automating responses, handling routine inquiries, and enabling rapid, personalized interaction. Businesses report conversion rates up to three times higher than traditional methods, with some sectors achieving 70% conversion and sales growth of 67% [10], [11], [12]. Beyond efficiency, emotionally adaptive chatbots, capable of adjusting tone, word choice, and even simulated cues, enhance satisfaction and loyalty [13], [14].

### D. Prominent Datasets for Intelligent Sales Dialogue Systems

Conversational shopping assistants depend on large, annotated corpora that teach models to capture user intents, clarify needs, and retrieve products [15]. Existing resources differ in collection methodology:

- Human–human dialogues: MultiWOZ (Wizard-of-Oz via Mechanical Turk) offers multi-domain conversations with goals, belief states, and dialogue acts [16]. MG-ShopDial provides coached e-commerce dialogues with utterance-level intent and goal labels [17]. PerSHOP includes 122 crowd-sourced dialogues annotated through rule-based and manual methods [18].

- Automated generation: SalesBot consists of LLM-generated sales dialogues annotated with user intents [19], reducing reliance on manual collection.

- Real customer-service exchanges: The Customer Support on Twitter dataset contains anonymized consumer-brand interactions via the Twitter API, while Bitext corpora supply specialized training material for retail and support chatbots [17].

High-quality annotation remains critical for supervised dialogue systems, yet manual collection is costly. Recent work shows LLMs can provide annotations at far lower expense [20]. To address these challenges, we propose a scalable multi-agent architecture for automatic question–answer dataset generation, minimizing manual intervention.

## E. Comparison with Existing Datasets

To position MegaChat within the sales dialogue landscape, we compare it against MultiWOZ, SalesBot 2.0, Customer Support on Twitter, MG-ShopDial, Bitext Retail/Support, and PerSHOP. TABLE I. summarizes release year, language coverage, and methodology. While each has advanced dialogue system research, MegaChat introduces distinct advantages:

- Fully Automated and Scalable: Unlike human-constructed datasets such as MultiWOZ and PerSHOP[17],[21], MegaChat is LLM-based and requires no human review, enabling rapid expansion across domains and languages. Automation may introduce subtle biases, which merit future study.

- Multi-Model Answer Generation with LLM Evaluation: Prior datasets provide static annotations. MegaChat integrates retrieval-augmented generation (GPT-4 variants) and an agentic LLM–SLM pipeline, with GPT-5.1 evaluating candidate answers to select ground truth—an absent feature in earlier corpora.

- Persona-Driven Question Generation: SalesBot 2.0 and Bitext corpora include synthetic dialogues but lack persistent persona realism[18], [20]. MegaChat models authentic user profiles with informal grammar and natural typing errors, yielding more realistic interactions.

- Two-Pass Refinement for Quality Control: Unlike datasets relying on initial annotation (e.g., MultiWOZ 2.2), MegaChat applies automated validation and refinement, filtering low-confidence outputs and enhancing conversational naturalness for accuracy and diversity.

## II. DATASET ANALYSIS AND PREPARATION

### A. Data Collection from Telegram Channels

We constructed the MegaChat dataset by retrieving the latest 5,000 posts from 48 active Persian Telegram shop channels across diverse e-commerce domains (e.g., clothing, electronics, home appliances, lifestyle). Non-textual or deleted posts were excluded(Fig. 4). The distribution of retained posts is shown in Fig. 1, highlighting differences in posting volume across categories.

### B. Temporal Posting and Engagement Benchmarks

To evaluate posting behavior and engagement patterns, we constructed two normalized benchmarks:

- Views by posting hour (Fig. 2): average views per post, normalized within each channel, then aggregated across all 48 channels. This highlights when posts tend to attract higher engagement.

- Posting activity by hour (Fig. 3): share of posts published in each hour, normalized per channel and averaged across all channels. This shows when shops are most active in posting.

Together, these benchmarks reveal both when shops post and when audiences engage most.

TABLE I. COMPARATIVE CHARACTERISTICS OF PROMINENT SALES AND SUPPORT DATASETS

| Dataset Name | Year | Language | Data Collection Method |
|---|---|---|---|
| Customer Support Twitter [17] | 2017 | English | Real-world Data (Automated scraping) |
| MultiWOZ 2.2 [16] | 2020 | English | Human-Human (Crowdsourcing) |
| SalesBot 2.0 [19] | 2023 | English | Fully Synthetic (LLM-generated) |
| MG-ShopDial [15] | 2023 | English | Coached Human-Human |
| Bitext Retail E-commerce [21] | 2024 | English | Hybrid Synthetic (NLG+NLP with human-in-the-loop) |
| Bitext Customer Support [21] | 2024 | English | Hybrid Synthetic (NLG+NLP with human-in-the-loop) |
| PerSHOP [18] | 2024 | Persian | Crowdsourcing + Manual Annotation |
| **MegaChat (ours)** | 2025 | Persian | Fully Synthetic (LLM-Agent-generated) |

### C. Content Markers and Audience Engagement

We analyzed two complementary aspects of the dataset:

- Hashtag usage (Fig. 5): counts of hashtags embedded in posts, reflecting how channels label and thematically organize their content.

- Emoji reactions (Fig. 6): counts of audience reactions by emoji type, showing how users most commonly respond to posts.

Together, these measures highlight both producer-side content markers and consumer-side engagement signals.

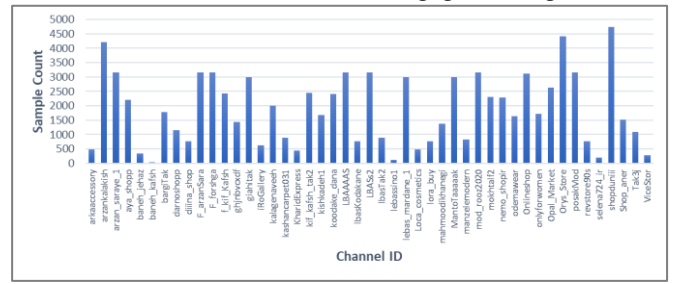

Fig. 1. Distribution of collected posts across 48 Telegram channels after filtering

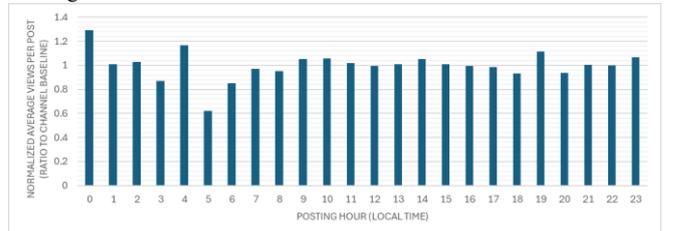

Fig. 2. Normalized average views per post by posting hour, aggregated across 48 channels (ratio to channel baseline).

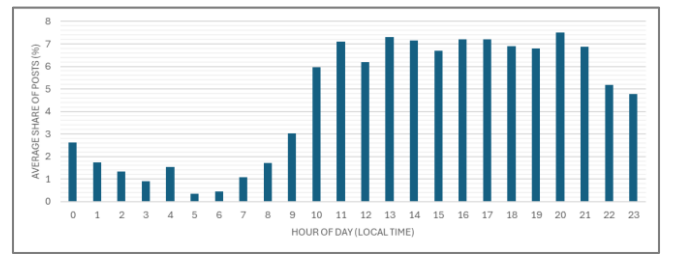

Fig. 3. Hourly posting activity: mean percentage of posts per hour, normalized per channel and averaged across all channels.

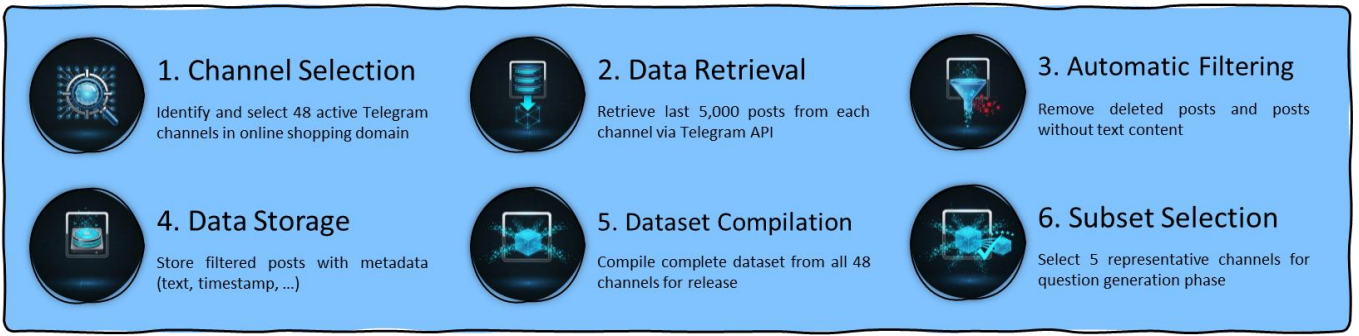

Fig. 4. Data Collection and Preparation Architecture

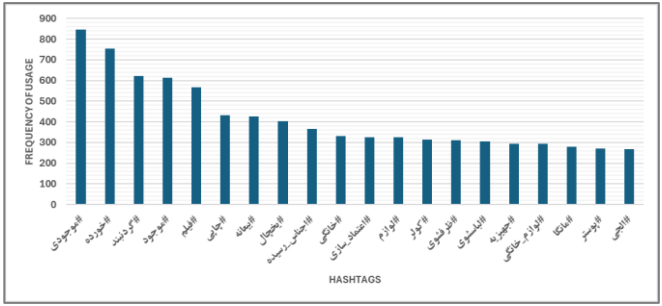

Fig. 5. Frequency of hashtag usage across posts, showing how channels label and organize content.

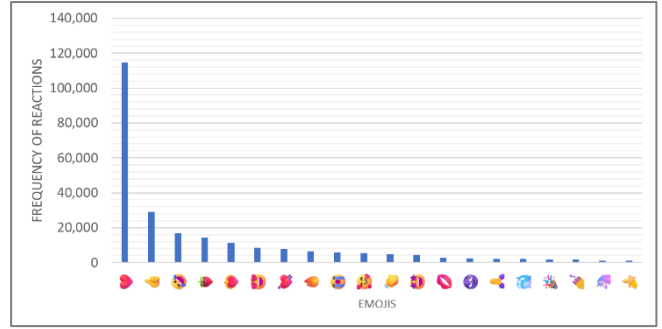

Fig. 6. Frequency of audience reactions by emoji, highlighting common engagement patterns with posts.

### D. Product Category Distribution

Channels were grouped into major product clusters: clothing (45.8%), home/appliances (12.5%), accessories/miscellaneous (~20%), and niche categories such as pet supplies, toys, and lifestyle items. Imports and luxury goods form a notable segment. Because some shops span multiple categories, total exceeds 100%. The distribution is summarized in TABLE II. .

## III. MEGACHAT DATASET DESCRIPTION

### A. Representative Channel Selection

From 48 Persian Telegram shop channels, we selected five that capture diverse e-commerce domains and communication styles. This subset ensures coverage across product categories, demographics, and linguistic registers, forming a solid basis for dataset generation.

The selected channels are:

- @LBASs2 (3,148 posts): Men's and women's clothing
- @nemo_shopir (2,295 posts): Anime and Pinterest-themed products, action figures, accessories, and manga
- @bargiTak (1,779 posts): Personal and home electrical appliances
- @mahmoodikhanegi (1,375 posts): Home appliances and bridal dowry items including refrigerators, washing machines, dishwashers, TVs, and air conditioners
- @lbasTak2 (881 posts): Children's clothing and apparel

This curated selection balances high-volume channels with niche domains, ensuring that subsequent sampling and synthetic question generation are both diverse and representative of real e-commerce dialogue contexts.

### B. Multi-Agent Question Generation

We employed a multi-agent architecture to generate realistic, persona-based questions in Persian for Telegram shopping channels. The system utilizes three specialized agents in a Two-Pass refinement pipeline:

Agent Roles:

- Generator Agent: Produces initial questions aligned with user personas and channel characteristics
- Validator Agent: Verifies question authenticity against source data
- Refiner Agent: Enhances conversational naturalness and filters low-confidence outputs

TABLE II. DISTRIBUTION OF PRODUCT CATEGORIES ACROSS CHANNELS, HIGHLIGHTING MAJOR CLUSTERS AND NICHE SEGMENTS.

| Cluster / Category Group | Count | % of Total (48) |
|---|---|---|
| 👕 Clothing (generic + men + women + kids + mixed) | 22 | 45.8% |
| 👜 Bags & Shoes (all variations) | 4 | 8.3% |
| ⌚ Accessories | 5 | 10.4% |
| 🏠 Home & Appliances | 6 | 12.5% |
| 🪴 Lifestyle & Decor | 3 | 6.3% |
| 🧸 Kids & Toys | 2 | 4.2% |
| 🐶 Pet Supplies | 1 | 2.1% |
| 🌸 Beauty | 1 | 2.1% |
| 🌍 Imports & Luxury | 3 | 6.3% |
| ❓ Miscellaneous | 5 | 10.4% |

Two-Pass Pipeline:

- Pass 1 (Generation): The generator combines channel metadata, sampled posts, and personas to produce colloquial Persian questions with informal grammar and assigns confidence scores.
- Pass 2 (Validation & Refinement): The validator removes unsupported claims, while the refiner enhances naturalness and filters questions below 50% confidence

Key Design Principles:

- Persona-driven generation ensuring questions reflect authentic user motivations
- Data-grounded outputs verified against actual channel content
- Conversational authenticity incorporating informal language and natural errors
- Confidence-based filtering maintaining high-quality standards

This multi-agent approach ensures diverse, contextually appropriate questions suitable for downstream tasks including chatbot training and user behavior analysis.

Fig. 7 presents the distribution of sampled posts and generated questions across the five Telegram shopping channels.

Fig. 8 illustrates the multi-agent pipeline for question generation, showing the flow from input sources through sequential agent processing to the final output.

### C. Answer Generation via Multi-Agent Architecture

Fig. 9 illustrates our answer generation pipeline, which consists of two parallel approaches: a Classic RAG system using three GPT-4 variants, and an Agentic System with a five-stage workflow combining LLMs and SLMs. All four systems generate candidate answers that are subsequently evaluated by GPT-5.1 to establish ground truth labels for the benchmark dataset.

#### 1) Classic RAG Implementation

The RAG baseline consists of a retriever (FAISS vector store with embeddings from OpenAI's text-embedding-3-large) and an LLM synthesizer. For each query, the retriever selects the top-5 posts via cosine similarity. The LLM prompt enforces:

- Strict grounding: Responses must be derived exclusively from retrieved channel posts, with explicit acknowledgment when information is unavailable
- Tone adaptation: The system matches the linguistic style and formality level of the user's query
- Operational detail inclusion: Responses include specific actionable information such as prices, Telegram handles, delivery timeframes, and product specifications when available

We evaluated three state-of-the-art LLMs to assess model-specific performance variations: GPT-4.1, GPT-4o, and GPT-4-turbo. Each model processed identical query-persona-context triplets, enabling direct comparison of response quality across different architectures.

#### 2) Agentic Architecture with LLM-SLM Collaboration

To overcome single-pass retrieval limits, we designed a five-stage agentic workflow combining LLMs for reasoning with SLMs for lightweight tasks:

- Query Generation (SLM): 5–8 diverse queries covering specs, pricing, delivery, alternatives
- Parallel Retrieval: concurrent search across all queries, aggregated into a candidate pool
- Re-ranking (SLM): scoring and selecting the top-5 most relevant posts
- User Profile Analysis (LLM): extracting persona preferences, emotional state, and style cues
- Answer Synthesis (LLM): generating responses that balance factual accuracy with persona-aligned tone

Advantages over Classic RAG: broader retrieval coverage, improved relevance, personalized communication, efficient division of labor between LLMs and SLMs, and higher overall response quality.

#### 3) Ground Truth Selection

To establish ground truth labels for our benchmark dataset, we implemented an LLM-as-a-judge evaluation framework using GPT-5.1. For each question in the dataset, we generated four candidate answers: three from classic RAG systems (GPT-4.1, GPT-4o, GPT-4-turbo) and one from the agentic architecture. The evaluator then ranks these answers to identify the best response as the ground truth label.

##### a) Evaluation Input

For each question $q$, the evaluator receives a tuple:

$$I(q) = (Q, P, U, A)$$

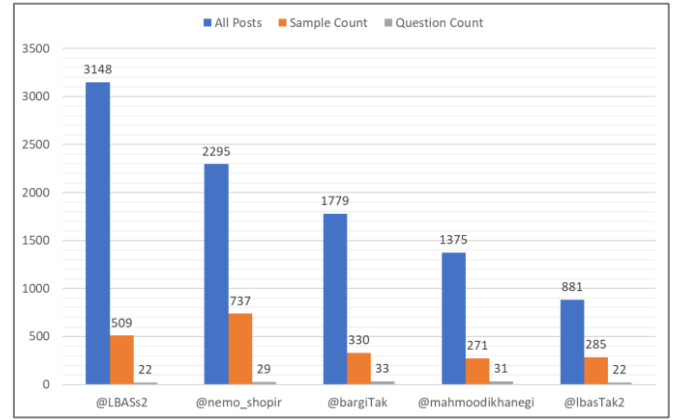

Fig. 7. Distribution of Sampled Posts and Generated Questions per Channel

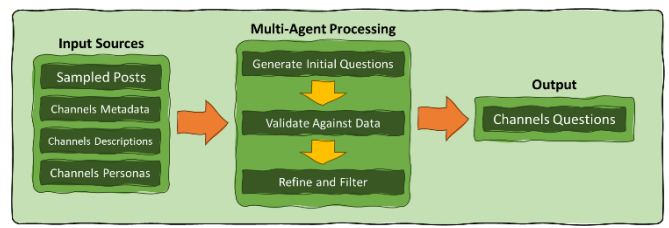

Fig. 8. Multi-Agent Architecture for Question Generation Pipeline

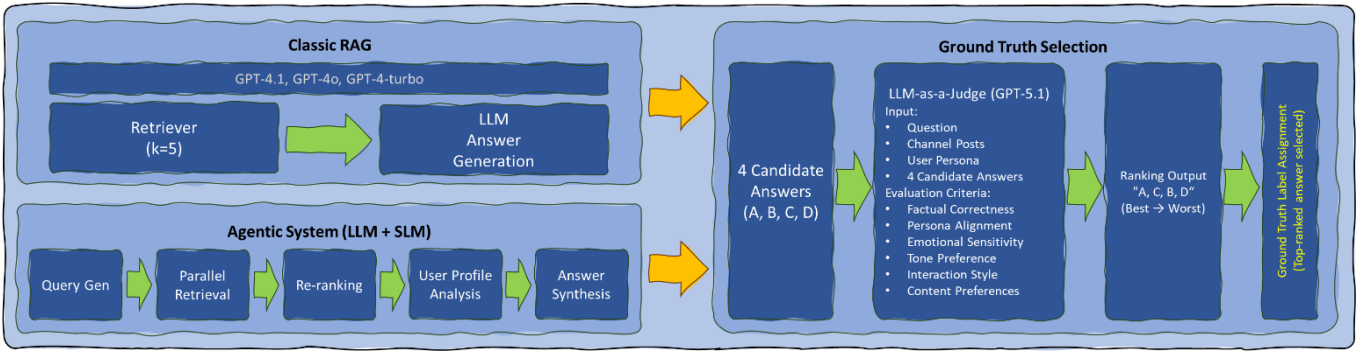

Fig. 9. Answer Generation and Ground Truth Selection Pipeline

Where:

- Q: User question in Persian
- P: Complete set of relevant Telegram channel posts
- U: User persona (demographics, psychographics, purchase behavior)
- A: Set of candidate answers, defined as

$$A = \{a_1, a_2, a_3, a_4\}$$

With each $a_i$ representing one candidate answer from generation systems.

*b) Ranking Criteria*

The evaluator considers six dimensions when ranking answers, with the following order of importance:

- Factual Correctness (FC): Strict grounding in posts, no hallucinations, explicit acknowledgment of missing information
- Persona Alignment (PA): Addressing user needs, motivations, and characteristics
- Emotional Sensitivity (ES): Appropriate empathetic or enthusiastic tone
- Tone Preference (TP): Formal, friendly, or professional style matching persona
- Interaction Style (IS): Clarity, conciseness, and preferred format (bullet points vs. paragraphs)
- Content Preferences (CP): Focus on relevant purchase motivations (price, discounts, warranties)

*c) Evaluation Function*

The evaluator applies a comparative ranking function across all criteria:

$$R: I(q) \rightarrow \sigma(q)$$

Where $\sigma(q)$ represents a permutation of $\{a_1, a_2, a_3, a_4\}$ ordered from best to worst based on holistic assessment of the six dimensions.

For any two answers $a_i$ And $a_j$:

$$a_i \succ a_j$$

$\leftrightarrow R$ evaluates $a_i$ as superior to $a_j$ across criteria $C = \{FC, PA, ES, TP, IS, CP\}$

The complete ranking is expressed as:

$$\sigma(q) = \langle a_{\pi(1)}, a_{\pi(2)}, a_{\pi(3)}, a_{\pi(4)} \rangle$$

Where $\pi$ is a permutation function such that:

$$a_{\pi(i)} \succ a_{\pi(j)} \text{ for all } i < j$$

*d) Ground Truth Selection*

The ground truth label for question $q$ is defined as:

$$GT(q) = \sigma(q)[1] = a_{\pi(1)}$$

representing the top-ranked answer in the evaluation output.

*e) Output Format*

The evaluator produces a ranking string in the format: "$a_{\pi(1)}, a_{\pi(2)}, a_{\pi(3)}, a_{\pi(4)}$" (e.g., "A, C, B, D"), where the leftmost answer is the ground truth and the rightmost is the lowest-ranked response.

IV. RESULTS AND DISCUSSION

The evaluation results across five Telegram shopping channels demonstrate the performance of four answer generation approaches: three classic RAG models (GPT-4.1, GPT-4o, GPT-4-turbo) and our proposed multi-agent architecture. The agentic architecture outperformed the RAG models in 4 out of 5 channels (@nemo_shopir, @bargiTak, @mahmoodikhanegi, and @lbasTak2), generating 8, 22, 18, and 15 best-ranked responses, respectively. This highlights the effectiveness of persona-driven analysis and multi-query retrieval. A plot of these results is shown in Fig. 10.

Additionally, a representative example of the generated question-answer pairs is illustrated in Fig. 11.

Despite some individual components underperforming, the GPT-5.1 evaluation layer ensured the selection of optimal answers, maintaining high dataset quality. These results demonstrate that a multi-agent system can create high-quality, persona-aware Q&A datasets without the need for complex fine-tuning or reinforcement learning, enabling scalable, automated generation across diverse domains and low-resource languages.

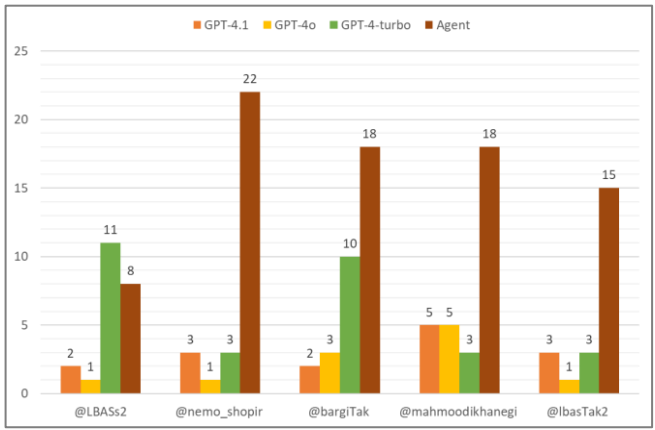

Fig. 10. Distribution of Best-Ranked Responses by Channel and Method

[Figure showing a Persian question-answer pair in a yellow box]

Fig. 11. Sample Question-Answer Pair from MegaChat Dataset

## V. CONCLUSION

This work introduced MegaChat, the first synthetic Persian Q&A dataset tailored for evaluating intelligent sales chatbots in Telegram-based e-commerce. By leveraging a fully automated, persona-aware multi-agent architecture, MegaChat demonstrates that scalable, high-quality datasets can be created for low-resource languages without costly human annotation or fine-tuning. Our findings highlight the dataset's novelty and practical impact: enabling SMEs to develop more effective customer engagement systems at reduced data acquisition costs, while advancing research in multilingual conversational AI and commercial domain benchmarking.

ACKNOWLEDGMENT

We would like to express our sincere gratitude to our families for their unwavering support and encouragement, to the faculty members at the universities for their valuable guidance throughout this research, and to colleagues and friends whose insights and discussions contributed to the development of this work.